\pdfoutput=1

\documentclass[11pt]{article}

\usepackage[acceptedWithA]{tacl2021v1}
\usepackage{booktabs}

\usepackage{times}
\usepackage{latexsym}
\usepackage{graphicx}
\usepackage{cleveref}
\usepackage{tikz}
\usepackage[edges]{forest}
\usepackage{makecell}
\usepackage{arydshln}

\usepackage[T1]{fontenc}
\usepackage[utf8]{inputenc}
\usepackage{enumitem}

\usepackage{pgfplots}
\definecolor{paired-light-blue}{RGB}{198, 219, 239}
\definecolor{paired-dark-blue}{RGB}{49, 130, 188}
\definecolor{paired-light-orange}{RGB}{251, 208, 162}
\definecolor{paired-dark-orange}{RGB}{230, 85, 12}
\definecolor{paired-light-green}{RGB}{149, 203, 211}
\definecolor{paired-dark-green}{RGB}{56, 177, 195}
\definecolor{paired-light-purple}{RGB}{218, 218, 235}
\definecolor{paired-dark-purple}{RGB}{117, 107, 176}
\definecolor{paired-light-gray}{RGB}{217, 217, 217}
\definecolor{paired-dark-gray}{RGB}{99, 99, 99}
\definecolor{paired-light-pink}{RGB}{222, 158, 214}
\definecolor{paired-dark-pink}{RGB}{123, 65, 115}
\definecolor{paired-light-red}{RGB}{238, 173, 152}
\definecolor{paired-dark-red}{RGB}{241, 131, 95}
\definecolor{paired-light-yellow}{RGB}{231, 204, 149}
\definecolor{paired-dark-yellow}{RGB}{141, 109, 49}
\definecolor{light-green}{RGB}{118, 207, 180}
\definecolor{raspberry}{RGB}{228, 24, 99}

\usepackage{microtype}
\usepackage{inconsolata}
\newcommand{\sparagraph}[1]{\noindent\textbf{#1.}}

\newcommand{\itparagraph}[1]{\noindent\textit{\underline{#1:}}}

\title{Culturally Aware and Adapted NLP: A Taxonomy and a Survey of the State of the Art}

\author{Chen Cecilia Liu\textsuperscript{1,2,3} \and Iryna Gurevych\textsuperscript{1} \and Anna Korhonen\textsuperscript{3}\\
\textsuperscript{1} UKP Lab,
Department of Computer Science and hessian.AI, \\
Technical University of Darmstadt\\
\textsuperscript{2} Konrad Zuse School of Excellence in Learning and Intelligent Systems (ELIZA)\\ 
\textsuperscript{3} Language Technology Lab, University of Cambridge\\
\texttt{\{chen.liu,iryna.gurevych\}@tu-darmstadt.de, alk23@cam.ac.uk}\\}

\begin{document}
\maketitle
\begin{abstract}
The surge of interest in \textit{culture} in NLP has inspired much recent research, but a shared understanding of ``culture'' remains unclear, making it difficult to evaluate progress in this emerging area. 
Drawing on prior research in NLP and related fields, we propose a fine-grained taxonomy of elements in culture that can provide a systematic framework for analyzing and understanding research progress. Using the taxonomy, we survey existing resources and methods for culturally aware and adapted NLP, providing an overview of the state of the art and the research gaps that still need to be filled.

\end{abstract}

\section{Introduction}\label{sec:intro}
\begin{figure}
    \centering
    \includegraphics[width=0.8\linewidth]{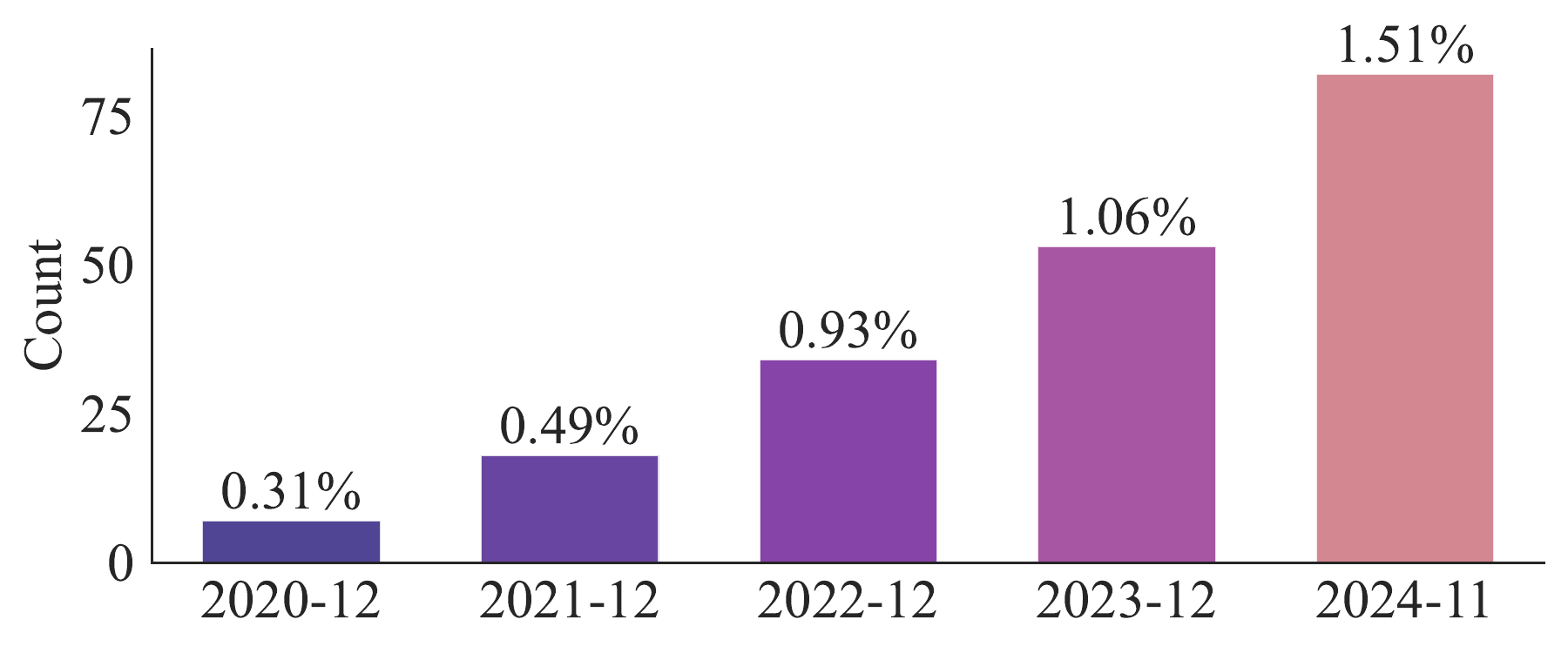}
    \caption{Papers title and abstract containing "culture" or "cultural" published in the main and findings of AACL/EACL/NAACL/ACL/EMNLP and TACL within 5 years, with normalized percentages based on the total number of papers at included venues to date.}
    \label{fig:growth}
\end{figure}

Culture is rapidly becoming an important research topic in Natural Language Processing (NLP), with a significant recent surge in the number of published papers (Figure~\ref{fig:growth}). 
Current NLP systems, especially Large Language Models (LLMs) often lack fairness and diversity in cultural awareness, which leads to biased performance that disproportionately favours certain groups, and causes harm to others \cite{DBLP:conf/fat/SambasivanAHDP21, DBLP:journals/corr/abs-2203-07785, analysis/cao-etal-2023-assessing, DBLP:journals/nature/HofmannKJK24}. To build technology that is equitable, inclusive and accessible, the NLP community must actively take the initiative, enhancing LLMs' cultural awareness and adaptability. 
Given the keen interest in this area and its importance for the safety and fairness of LLMs, it is now important to consolidate existing research on culturally aware and adapted NLP to take stock of the progress made so far and to identify research gaps. However, this is challenged by the lack of a common understanding of the concept of ``culture'' in NLP.

Prior work in NLP such as \citet{DBLP:conf/acl/HershcovichFLLA22} laid the vital foundations for understanding how language, culture and society interact. \citet{DBLP:conf/acl/HershcovichFLLA22} proposed a simple taxonomy derived from the interaction between language and culture that captures broad elements of culture (linguistic form and style, objectives and values, common ground, and aboutness). More recently, \citet{DBLP:journals/corr/abs-2403-15412} adopted ``proxies of culture'' (semantic or demographic proxies). While neither provides a shared understanding of culture, perhaps unsurprisingly, \textit{\textbf{language}} is an essential component of culture in NLP. 

\begin{figure*}[t!]
    \centering
    \includegraphics[width=\linewidth]{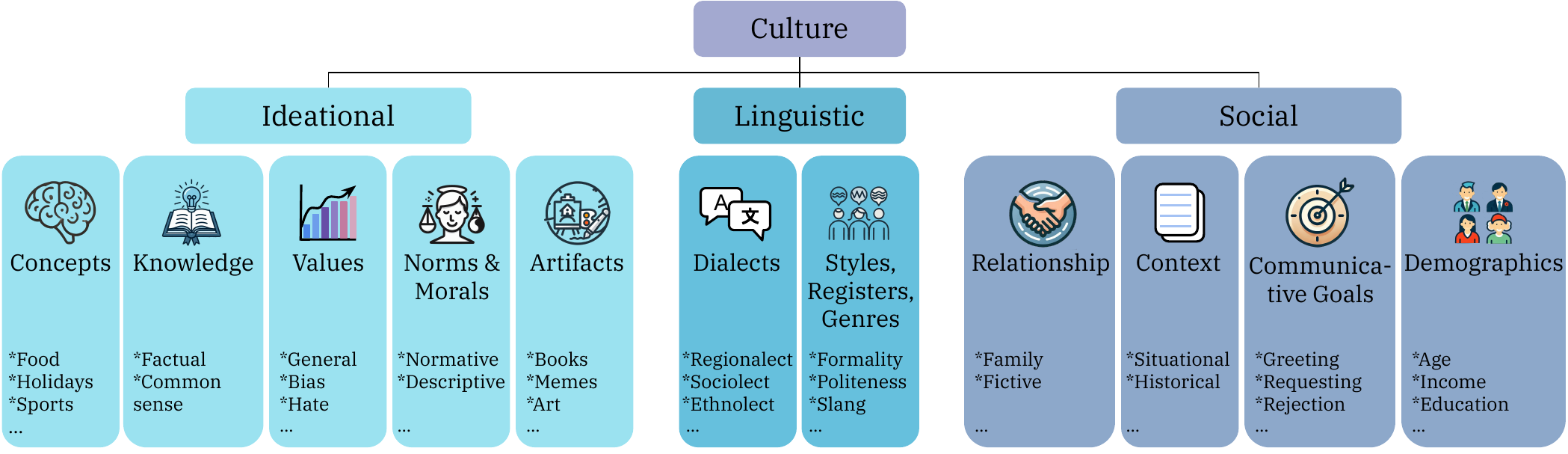}
    \caption{An overview of the taxonomy with examples of subcategories of future possible expansions. The elements in culture are organized into three different branches: \textbf{ideational}, \textbf{linguistic}, and \textbf{social}. The \textbf{ideational} branch (\S\ref{sec:cult_elements} encompasses the non-material aspects of culture that constitute a way of life. The \textbf{linguistic} branch (\S\ref{sec:forms}) focuses on cultural variations in language and linguistic forms, bridging the ideational and social elements of culture. The \textbf{social} branch (\S\ref{sec:sociocult}) covers key factors in social interaction and communication.}
    \label{fig:cat}
\end{figure*}

A shared understanding of culture in NLP could benefit from examining definitions developed in anthropology and social sciences.\footnote{They have been thinking about culture for over a century!}
In these fields~\citep{tylor1871primitive, kroeber1952culture, white1959concept, unesco1982world, matsumoto1996culture, culture/blake, geertzinterpretation, goffman2023presentation}, most definitions of culture involve \textit{people}, groups, and the interactions within and between individuals and groups.

\citet{murdock1940cross} describes culture as ``ideational''\footnote{The values, beliefs, norms, and ideas that comprise a way of life \cite{murdock1940cross,Briggle_Mitcham_2012}.} and ``social''. \citet{white1959concept} describes the \textit{locus of culture} as:  \textbf{1)} ``\textit{within human}'' (e.g., concepts, beliefs, i.e., \textbf{\textit{ideational}}), \textbf{2)} between ``\textit{social interaction} among human beings'', and \textbf{3)} outside of human but ``\textit{within the patterns of social interaction}''. 
An examination of such work reveals that \textit{\textbf{social} interactions} are critical components of culture, in addition to ideational elements. Notably, this social aspect is underrepresented in previous cross-cultural NLP research~\cite{DBLP:conf/acl/HershcovichFLLA22, DBLP:journals/corr/abs-2403-15412}.\footnote{There is also prior work focusing on social NLP~\cite{hovy-yang-2021-importance} which is related. However, culture is not the central theme of the work.} 

Hence, we combine the emphasis on \emph{language} in prior NLP work, while integrating the significance of \emph{social} and \emph{ideational} elements that shape culture. This leads us to a working definition of culture used to taxonomize recent work in NLP in this survey: \textit{Culture encompasses the collective ideas, shared language, and social practices that emerge from and evolve through human social interactions within a society}.

Grounded in this working definition, we introduce a new fine-grained taxonomy of culture by expanding on the basic categories of prior work~\cite{hovy-yang-2021-importance,DBLP:conf/acl/HershcovichFLLA22} to address above-mentioned issues. We then use this taxonomy to organize existing works in culturally aware and adapted NLP, and identify research gaps. Our survey of 127 publications in leading *CL venues (see selection method in Appendix~\ref{app:method}) provides an up-to-date view of cultural adaptation resources and models and identifies areas of progress as well as new research opportunities. We hope our taxonomy and analysis enable and inspire further research in this important emerging area.

\section{The Taxonomy}\label{sec:framwork}

In this section, we present our new taxonomy of culture.
Unlike previous NLP studies \cite{DBLP:conf/acl/HershcovichFLLA22, hovy-yang-2021-importance} that sought to define cultural elements, this taxonomy (i) is grounded in well-established elements of culture in anthropology and social sciences~\cite{tylor1871primitive, kroeber1952culture, white1959concept}, (ii) consists of more fine-grained elements than in earlier work, and (iii) allows for a wider consideration of how social factors and variations in humans influence culture.

Figure~\ref{fig:cat} presents a taxonomy of cultural elements derived from our working definition in \S\ref{sec:intro}, organized into three main branches: \textbf{ideational}, \textbf{linguistic}, and \textbf{social}.\footnote{Icons in Figure~\ref{fig:cat} are created with the assistance of DALL-E.} The \textbf{ideational} branch (\S\ref{sec:cult_elements}; \citealt{murdock1940cross, briggle2012ethics}) encompasses the non-material aspects of culture that constitute a way of life, such as values or knowledge. The \textbf{linguistic} branch (\S\ref{sec:forms}) focuses on cultural variations in language and linguistic forms, bridging the ideational and social elements of culture. The \textbf{social} branch (\S\ref{sec:sociocult}) covers key factors in social interaction and communication, such as relationships or communicative goals.\footnote{All cultural elements can interact and influence each other based on context and can be divided into finer groups. Similar to prior work, our taxonomy abstracts away from these contextual variations.} 
Here, we define each element based on existing research and relating to example tasks in the NLP context. We then provide details and examples from the current literature in \S\ref{sec:cult_elements}, \S\ref{sec:forms} and \S\ref{sec:sociocult}.

\noindent\textbf{{Ideational elements}} are based on well-established discussions of culture~\cite{tylor1871primitive, kroeber1952culture, white1959concept}:

\itparagraph{Concepts} basic units of meaning that structure and facilitate thought, bridging sensory experience~\cite{jackendoff1989concept, jackendoff2012concept}, e.g., cuisines (such as schnitzel, ratatouille) or holidays (such as Diwali, Nowruz). Related NLP task examples: question answering, dialogue generation.

\itparagraph{Knowledge} information that can be acquired through education or practical experience, e.g., local agricultural knowledge. Related NLP task examples: dialogue generation, reasoning.

\itparagraph{Values} beliefs, desirable end states or behaviours ranked by importance that can guide evaluations of things~\cite{value/definition/schwartz1992universals}. Unlike norms and morals, values do not inherently involve ethical judgment, e.g., beauty standards, or perception of hate speech. Related NLP task examples: content moderation, debiasing. 

\itparagraph{Norms and Morals} set of rules or principles that govern people's behaviour and everyday reasoning~\cite{cialdini1991focus, bicchieri2018social, hechter2001social, gert2002definition}, e.g., filial obedience attitude. Related NLP task examples: reasoning, safety alignment.   

\itparagraph{Artifacts} items that are products of human culture, such as art, poetry~\cite{white1959concept} etc. This is ideational in our taxonomy since we do not work on physical buildings or tools in NLP. Related NLP task examples: machine translation of long-form literature, emotion arc analysis of movies, memes classification.  

\noindent\textbf{{Linguistic elements}} relate to language variations in the cultural context, based on existing discussions \cite{lang_var/wardhaugh2021introduction}:  

\itparagraph{Dialects} includes variations of languages in a systematic way (\citealt{dialect_def/fromkin1998introduction, dialect_def/trudgill2000sociolinguistics, lang_var/wardhaugh2021introduction}; such as dialects continuum, regionalects, sociolects etc.), e.g., African American (Vernacular) English (AAE/AAVE). Related NLP task examples: machine translation, debiasing.  

\itparagraph{Styles, Registers, Genres} includes elements such as formality, variations of language in situation and communicative forms \cite{lang_var/wardhaugh2021introduction}, e.g., formality in text, slang, or specific genres like news, folk tales. Related NLP task examples: style transfer, creative writing generation.  

\noindent\textbf{{Social elements}} focus on social interactions and communication among humans within the scope of NLP. Leveraging the work of ~\citep{hovy-yang-2021-importance}, we identify relevant elements: 

\itparagraph{Relationship} connection between two or more individuals or groups, e.g., father-son, colleagues. Related NLP task examples: creative writing generation, dialogue generation.  

\itparagraph{Context} the ``containers'' of communications~\cite{yang2019computational}, which can be linguistic such as surrounding sentences or extra-linguistic~\cite{hovy-yang-2021-importance} including social settings (e.g., at a wedding), non-verbal cues (e.g., gesture), or historical contexts (e.g., colonization). Related NLP task examples: coreference resolution, pragmatic inference.

\itparagraph{Communicative Goals} the intention behind language use, e.g., requests, apologies, persuasion. Related NLP task examples: intent classification, emotion classification, human-AI collaboration.

\itparagraph{Demographics} the characteristics of people, e.g., economic income, education level, nationality, location, political view, family status, etc.  Related NLP task examples: content moderation, personalization.

\section{Elements of Culture in Current NLP (Resources) Literature}\label{sec:all_elements}

In this section, we survey and categorize NLP resources. Table~\ref{tab:cult_elements} shows an overview of papers organized according to the taxonomy.

We observe that in resources, culture can be captured in \textbf{1)} the data itself, or \textbf{2)} in the labels (e.g., multi-culturally annotated).  Further, while cultural differences are evident in linguistic and social elements, most current work relies on standard language or country boundaries, leaving these elements understudied.

\begin{table*}[h]
\centering
\resizebox{\textwidth}{!}{

    \begin{tabular}{l c}
    \toprule
    \textbf{Element} & \textbf{Papers} \\
    \midrule
     Concepts &  \makecell{
     \citealt{concept/time/acl/shwartz-2022-good,
     concept/adapt/tacl/MajewskaRPVK23,
     concept/multi3woz/tacl/Hu00609,
     concept/mabl/acl/KabraLKAWCAON23}\\ 
     \citealt{
     concept/maps/naacl/liu2023,
     adapt/ki/recipes/tacl_a_00634,
     concept/acl/jiang-joshi-2024-cpopqa,
     adapt/retrieval/emnlp/hu-etal-2024-bridging}\\
     \textbf{Vision-Language:} \citealt{
     concept/marvl/emnlp/0001BPRCE21,
     concept/gdvcr/emnlp/YinLHPC21,
     concept/xm3600/emnlp/ThapliyalPCS22}\\
     \citealt{
     concept/khanuja2024image,
     concept/foodi/emnlp/li-etal-2024-foodieqa,
     concept/emnlp/bhatia-etal-2024-local,
     concept/vqa/nayak-etal-2024-benchmarking}\\
     }\\
    \hdashline
    Knowledge &    
\makecell{\textbf{Probing:} \citealt{knowledge/mlama/eacl/kassner2021multilingual, knowledge/geomlama/emnlp/YinBMLC22, knowledge/dlama/keleg-magdy-2023-dlama, knowledge/FMLAMA/zhou2024does, emnlp/bhatt-diaz-2024-extrinsic}\\
    \textbf{MMLU:} \citealt{knowledge/indommlu/emnlp/KotoA0B23,  knowledge/cmmlu/li2023,knowledge/arabicmmlu/Koto2024ArabicMMLUAM,
    knowledge/seaeval/abs-2309-04766, knowledge/kmmlu/abs-2402-11548}\\ 
    \textbf{Common sense:} \citealt{knowledge/xcopa/ponti-etal-2020-xcopa, knowledge/copalid/abs-2311-01012, knowledge/indocult/koto2024indoculture,
    knowledge/emnlp/acquaye-etal-2024-susu,
    knowledge/culturebank/emnlp/shi-etal-2024-culturebank}\\
    }\\
    \hdashline
    Values - general  & \makecell{
    \citealt{value/acl/tay-etal-2020-rather,
    nvm/acl/ramezani-xu-2023-knowledge, 
    analysis/cao-etal-2023-assessing,
    values/thanksgiving/acl/wang2024}\\
    \citealt{
    nvm/naacl/FULCRA/yao-etal-2024-value, 
    adapt/rlhf/arash-etal-2024-multilingual}\\
% \citealt{DBLP:journals/corr/abs-2309-12342, 
% nvm/kirk2024prism, nvm/worldvaluesbench/zhao-etal-2024-worldvaluesbench-large} 
    }  \\
    \cdashline{2-2}
    Values - bias &   \makecell{
   \textbf{WEAT:} \citealt{
    bias/weat/naacl/malik-etal-2022-caste,
    bias/ca-weat/emnlp/espana-bonet-barron-cedeno-2022-undesired, 
    bias/weathub/mukherjee-etal-2023-global,
        bias/chbias/acl/ZhaoFSL0P23}\\
    \citealt{
    bias/hostility/acl/SahooMB23,
    bias/naacl/camel/abs-2305-14456,
    bias/seegull/jha-etal-2023-seegull,
    bias/seegull/bhutani2024seegull,
    value/mukherjee-etal-2024-global}\\
    \textbf{Sent. Pairs:} \citealt{
    nangia-etal-2020-crows,
    bias/frcrowspairs/acl/NeveolDBF22,
    bias/winoqueer/acl/FelknerCJM23,
    value/naacl/sahoo-etal-2024-indibias}\\
    \textbf{Other:} \citealt{
    genderbias/campolungo-etal-2022-dibimt,
    genderbias/dnic/sandoval-etal-2023-rose,
     genderbias/attanasio-etal-2023-tale,
    bias/socialcommon/eacl/BauerTB23}\\
    \citealt{
    bias/CDIAL-BIAS/acl/zhou-etal-2022-towards-identifying,
    bias/kosbi/acl/lee-etal-2023-kosbi,
    bias/fork/acl/PaltaR23,
    bias/sodapop/eacl/an-etal-2023-sodapop,
    bias/kobbq/tacl/jin23,
    value/bias/taiwan/hsieh-etal-2024-twbias}\\
    }\\
      \cdashline{2-2}
    Values - hate &   \makecell{
    \citealt{
    nvm/aacl/coral/shekhar-etal-2022-coral,
    hate/eacl/zhou-etal-2023-cross,
    hate/cultcompass/emnlp/ZhouKCH23,
    bias/crehate/abs-2308-16705,
    hate/d3code/emnlp/mostafazadeh-davani-etal-2024-d3code}
    }\\ 
      \cdashline{2-2}
    Values - other perceptions &   \makecell{\citealt{
    art/artelingo/emnlp/MohamedAAL00E22,
    hate/frenda-etal-2023-epic, 
    hate/casola-etal-2024-multipico,
    concept/lexica/havaldar-etal-2024-building}\\
    \citealt{
    art/artelingo/emnlp/mohamed-etal-2024-culture,
    emotion/emnlp/deas-etal-2024-masive}\\
    }\\
    \hdashline
    Norms and Morals  & \makecell{ 
    \citealt{
    nvm/socchem/emlnp/forbes-etal-2020-social,
    nvm/emnlp/emelin-etal-2021-moral,
    nvm/mic/acl/ziems-etal-2022-moral,
    nvm/prosocial/emnlp/kim-etal-2022-prosocialdialog,
    nvm/normmark/acl/moghimifar-etal-2023-normmark}\\
    \citealt{
    nvm/normsage/emnlp/fung-etal-2023-normsage,
    nvm/clarifydelphi/acl/pyatkin-etal-2023-clarifydelphi,
    nvm/SocNormNLI/emnlp/ch-wang-etal-2023-sociocultural,
    nvm/normbank/acl/ziems-etal-2023-normbank, 
    nvm/eticor/emnlp/dwivedi-etal-2023-eticor}\\
    \citealt{
    nvm/deltaROT/acl/kavel-etal-2023,
    nvm/culturallyawareNLI/huang-yang-2023-culturally,
    nvm/moraldial/acl/sun-etal-2023-moraldial, 
    nvm/normdial/emnlp/li-etal-2023-normdial}\\
    \citealt{
    nvm/renovi/naacl/abs-2402-11178,
    nvm/acl/kim-etal-2024-moralemotion,
    emnlp/bhatt-diaz-2024-extrinsic,
    adapt/ki/vijjini-etal-2024-socialgaze,
    nvm/emnlp/liu-etal-2024-evaluating-moral}\\
    }\\
    \hdashline
    Artifacts &  \makecell{ \citealt{
    art/emnlp/music/epure-etal-2020-modeling,
    art/artelingo/emnlp/MohamedAAL00E22,
    art/emnlp/kruk-etal-2023-impressions}\\
    \citealt{
    art/jiang-etal-2023-discourse,
    art/artelingo/emnlp/mohamed-etal-2024-culture,
    art/emnlp/hobson-etal-2024-story}\\
    }\\
    \midrule
    \midrule
    Dialects & \makecell{ 
    \citealt{
    dialect/value/acl/ziems-etal-2022-value,
    dialect/multivalue/acl/ziems-etal-2023-multi,
    dialect/viet/emnlp/le-luu-2023-parallel,
    dialect/swiss/acl/pluss-etal-2023-stt4sg,
    dialect/dialect2std/emnlp/kuparinen-etal-2023-dialect}\\ 
    \citealt{
    dialect/elmadany-etal-2023-orca,
    dialect/aave/emnlp/deas-etal-2023-evaluation,
    dialect/khondaker-etal-2023-gptaraeval,
    dialect/dialectbench/faisal2024dialectbench}\\
    }\\
    \hdashline
    Styles, Registers, Genres & \makecell{ 
    \citealt{
    slang/naacl/sweed-shahaf-2021-catchphrase,
    style/emnlp/sun-xu-2022-tracing,
    formality/naacl/nadejde-etal-2022-cocoa}\\
    \citealt{
    politness/emnlp/srinivasan-choi-2022-tydip,
    politness/emnlp/havaldar-etal-2023-comparing}\\
    }\\
    \midrule\midrule
    Relationship  & \makecell{ 
    \citealt{
    nvm/normdial/emnlp/li-etal-2023-normdial,
    sociocult/relation/jurgens-etal-2023-spouse,
    adapt/ki/codenames/shaikh-etal-2023-modeling,
    nvm/normbank/acl/ziems-etal-2023-normbank,
    nvm/renovi/naacl/abs-2402-11178}
    }\\
    \hdashline
    Context & \makecell{ 
    \citealt{
    nvm/socchem/emlnp/forbes-etal-2020-social,
    nvm/emnlp/emelin-etal-2021-moral,
    nvm/prosocial/emnlp/kim-etal-2022-prosocialdialog,
    nvm/normbank/acl/ziems-etal-2023-normbank,
    nvm/normmark/acl/moghimifar-etal-2023-normmark}\\
    \citealt{
    nvm/deltaROT/acl/kavel-etal-2023,
    nvm/moraldial/acl/sun-etal-2023-moraldial,
    nvm/SocNormNLI/emnlp/ch-wang-etal-2023-sociocultural,
    nvm/renovi/naacl/abs-2402-11178}\\
    }\\
        \hdashline
    Communicative Goals & \makecell{
    \citealt{nvm/emnlp/emelin-etal-2021-moral,
    nvm/normdial/emnlp/li-etal-2023-normdial,
    nvm/renovi/naacl/abs-2402-11178}\\
    }\\
        \hdashline
    Demographics & \makecell{ 
    \citealt{hate/frenda-etal-2023-epic, 
    demographics/emnlp/lahoti-etal-2023-improving,
    nvm/normbank/acl/ziems-etal-2023-normbank,
    hate/casola-etal-2024-multipico,
    bias/crehate/abs-2308-16705}
    }\\
    \bottomrule
    \end{tabular}
    }
    \caption{Recent resource work considered in \S\ref{sec:all_elements} by elements (selection method in Appendix~\ref{app:method}). The three blocks (divided by double lines) correspond to ideational, linguistic, and social elements respectively.}
    \label{tab:cult_elements}
\end{table*}

\subsection{Ideational Elements}\label{sec:cult_elements}

\subsubsection{Concepts}\label{sec:concept}
We can divide concepts into 1) basic concepts that are ``configured'' differently, reflecting the cultural-specific way of thinking,\footnote{For example, one can explore the citizen science project for lexicon associations: \url{https://smallworldofwords.org/en/project/home}.} 
and 2) concepts that are unique to a culture~\cite{wierzbicka1992semantics}.\footnote{For example, ``Kopi Ga Dai'' in Singaporean English versus ``double-double'' in Canada, both referring to coffee with extra sweetness and creaminess, but very different.} 

Recent NLP research has explored grounding time expressions across cultures \citep{concept/time/acl/shwartz-2022-good} and culinary concepts in recipe adaptations \citep{adapt/ki/recipes/tacl_a_00634}. Additionally, studies have examined how various cultures \emph{use} concepts across categories, such as through metaphors \cite{concept/mabl/acl/KabraLKAWCAON23} or traditional proverbs and sayings \cite{concept/maps/naacl/liu2023}. In vision and language (VL) settings, culturally unique concepts have been integrated into reasoning and captioning tasks \citep{concept/marvl/emnlp/0001BPRCE21, concept/gdvcr/emnlp/YinLHPC21, concept/xm3600/emnlp/ThapliyalPCS22, concept/foodi/emnlp/li-etal-2024-foodieqa} or assess multimodal content adaptations \cite{concept/khanuja2024image} and generation of text-to-image models \cite{DBLP:journals/corr/abs-2301-12073}.

These datasets are often small due to high annotation costs, and most are only available for evaluation. Training and evaluation datasets still lack diversity across cultures, languages, and concept categories (e.g., rituals, aesthetics, spatial relations).

\subsubsection{Knowledge}\label{sec:knoweldge}

Cultural knowledge can be factual or common sense.\footnote{Common sense and norms are sometimes used interchangeably in NLP. Norms are acceptable behavioural patterns of a group (\S\ref{sec:framwork}), which we will discuss in \S\ref{sec:nvm}.} \textit{What weather phenomena can be expected if a rapidly rotating tropical storm forms off the coast of our country?} (It's likely called a hurricane if one is in US, a typhoon if one is in Korea.) \textit{Is tofu pudding sweet or salty by default?} (In China, it's typically sweet in the south but salty in the north.)

We identified three major types of resources in NLP literature: 
1) probing (by masking entities), 2) multiple choice question answering (MCQA), and 3) knowledge bases.

Assessing language models' knowledge has long been important, with early studies examining this across languages predating LLMs~\cite{knowledge/mlama/eacl/kassner2021multilingual, knowledge/geomlama/emnlp/YinBMLC22, knowledge/dlama/keleg-magdy-2023-dlama, knowledge/FMLAMA/zhou2024does}. Recently, MMLU-style~\cite{knowledge/mmlu/iclr/HendrycksBBZMSS21} MCQA benchmarks have advanced LLM development and inspired cultural variants (e.g., \citealt{knowledge/cmmlu/li2023}, details in Table~\ref{tab:cult_elements}) covering aspects like food, history, and geography in respective languages. However, MMLU-style benchmarks, often based on standard exams and textbooks, lack integration with broader cultural elements. In contrast, other common sense knowledge datasets (e.g., \citealt{knowledge/copalid/abs-2311-01012}, \citealt{knowledge/indocult/koto2024indoculture}) can incorporate other elements under ``linguistic'' (e.g., in \textit{dialects}) or ``social'' (such as from diverse \textit{demographics} with geographic regions) branches.

Finally, integrating knowledge bases (KB) with models enhances cultural awareness \cite{adapt/cpt/emnlp/bhatia-shwartz-2023-gd} and supports culturally relevant synthetic data generation \cite{knowledge/soda/kim-etal-2023-soda}. Despite recent efforts in creating cultural KB from other venues \cite{kb/www/candle/NguyenRVW23, kb/culturalatlas/fung2024massively, kb/mango/abs-2402-10689}, *CL community examples remain limited.

\subsubsection{Values}\label{sec:values}

Diverse ranking of values among groups can result in differences in aboutness, communication styles, perceptions and multiple other dimensions~\cite{hofstede1984culture, value/hofstede2011dimensionalizing}. Such differences in pre-training data can be reflected in LLMs.  

Many recent studies on evaluation~\cite[inter alia]{DBLP:journals/corr/abs-2203-07785, nvm/acl/ramezani-xu-2023-knowledge, analysis/cao-etal-2023-assessing, GlobalOpinionQA/corr/abs-2306-16388, demographic/icml/opinionsQA/SanturkarDLLLH23, DBLP:journals/corr/abs-2309-12342, value/wassa/HavaldarSRLGU23, values/thanksgiving/acl/wang2024} show that LLMs align better with values of WEIRD (Western, Educated, Industrialized, Rich and Democratic,~\citealt{henrich2010weirdest}) people, raising concerns about the fairness and safety of LLMs for others. Here, Pew Global Attitudes Survey (PEW)\footnote{\url{https://www.pewresearch.org/}}, the World Values Survey (WVS)\footnote{\url{https://www.worldvaluessurvey.org/}} and the Hofstede Cultural Dimensions~\cite{hofstede1984culture, value/hofstede2011dimensionalizing} are commonly used for evaluation, along with regional variants like the European Values Survey (EVS, \citealt{nvm/evs/ZA4438}). However, the questions of how to improve the model's value alignment with diverse cultures, what resources to collect and whom to collect from remain unsolved~\cite{nvm/kirk2024prism}.

\sparagraph{Biases} 
In contrast to general cultural values, biases have been long-studied in NLP, such as gender bias in machine translation~\cite[inter alia]{genderbias/acl/stanovsky-etal-2019-evaluating, genderbias/tacl_a_00401,genderbias/campolungo-etal-2022-dibimt,genderbias/dnic/sandoval-etal-2023-rose,genderbias/attanasio-etal-2023-tale} or bias towards particular social groups. 
Differences in value ``ranking'' lead cultures to exhibit distinct biases toward the same groups or unique biases specific to certain cultures (e.g., caste systems, unnatural beauty standards). These variations are central to the study of \emph{cultural biases} and are the focus of our work.

To enable evaluations of cross-cultural variations in biases and develop transferable de-biasing methods, recent work has created varieties of culturally aware datasets to aid evaluations, including targets and attribute word sets, sentence pairs, conversational and QA data (see Table \ref{tab:cult_elements} for the papers).

Overall, this area shows notable progress compared to other sub-areas. Recent surveys on general biases cover key topics like evaluation and de-biasing methods~\cite{bias/survey/sun-etal-2019-mitigating, bias/survey/meade-etal-2022-empirical, bias/survey/dev-etal-2022-measures, bias/survey/delobelle-etal-2022-measuring}, which we refer readers to them for further details.

\sparagraph{Hate} Like biases, perceptions of hatefulness in the text also vary across cultures, as shown in recent research on hate speech classification~\citep{demo/naacl/SapSVZCS22, hate/eacl/zhou-etal-2023-cross, hate/cultcompass/emnlp/ZhouKCH23, hate/lee-etal-2023-hate, hate/GeoOLID/coling/lwowski-etal-2022-measuring, hate/arango-monnar-etal-2022-resources}. Such model disparities may be due to the data source (i.e., using machine translations, not native text) or the labelling process. The first issue can be addressed by diversifying data sources, incorporating authentic local data \citep{nvm/aacl/coral/shekhar-etal-2022-coral, hate/emnlp/jeong-etal-2022-kold}. The second issue can be mitigated by creating annotations from diverse cultural groups. Recently, CREHate~\cite{bias/crehate/abs-2308-16705} investigates variations in hate speech perceptions within the same language, highlighting the need for further research.

\sparagraph{Other Perceptions}
The perception of politeness, aesthetic appeal or emotions can also vary across cultures~\cite{perception/politness, mesquita11997culture, masuda2008culture, perception/sarcasm/ringel-etal-2019-cross, value/depression/abdelkadir-etal-2024-diverse}. 
For example, whether a piece of text is deemed humorous or ironic is culturally dependent. \citet{hate/frenda-etal-2023-epic, hate/casola-etal-2024-multipico} try to address this with cross-cultural annotated (multilingual) irony corpora. Similarly, visual elements in arts can elicit different emotions in different cultural groups. ArtELingo~\cite{art/artelingo/emnlp/MohamedAAL00E22,art/artelingo/emnlp/mohamed-etal-2024-culture} provides benchmarks with multilingual captions and emotion labels for artworks to evaluate models' cultural-transfer performance. This research area is significantly limited.

\subsubsection{Norms and Morals}\label{sec:nvm}

In ethics, a distinction is made between descriptive and normative morality~\cite{gert2002definition}. In NLP, this distinction is often overlooked~\cite{nvm/survey/vida-etal-2023-values} with a greater emphasis on the ``end product'', which is the final set of rules or principles and their judgments.\footnote{This is reasonable for standard NLP tasks but should be re-evaluated for high-stakes judgment-based applications.} 

Several norm banks exist, built through automatic, semi-automatic, or manual methods using sources like conversations, social media, or government websites~\cite{nvm/socchem/emlnp/forbes-etal-2020-social,nvm/normsage/emnlp/fung-etal-2023-normsage, nvm/SocNormNLI/emnlp/ch-wang-etal-2023-sociocultural, nvm/eticor/emnlp/dwivedi-etal-2023-eticor}. These norm banks have also been automatically adapted to defensible norms in fine-grained situations~\cite{nvm/clarifydelphi/acl/pyatkin-etal-2023-clarifydelphi, nvm/deltaROT/acl/kavel-etal-2023} or inference tasks~\cite{nvm/SocNormNLI/emnlp/ch-wang-etal-2023-sociocultural, nvm/culturallyawareNLI/huang-yang-2023-culturally} for LLM evaluation and adaptation.
 
For model alignment, several approaches focus on ``inquisition'', directly questioning LLMs about issues through conversation or a QA task \cite{nvm/prosocial/emnlp/kim-etal-2022-prosocialdialog, nvm/moraldial/acl/sun-etal-2023-moraldial,  nvm/acl/yu-etal-2024-cmoraleval,nvm/acl/lee-etal-2024-kornat, nvm/naacl/social/yuan-etal-2024-measuring}. A challenge with this approach is that a model's responses do not always align with its behaviour in usage (i.e., conversation). Thus, culturally aligned conversational data show greater potential for behaviour adaptation ~\cite{nvm/normdial/emnlp/li-etal-2023-normdial, nvm/renovi/naacl/abs-2402-11178}. However, existing resources have limited coverage beyond Western, Chinese, and Indian cultures.

\subsubsection{Artifacts}\label{sec:artifacts}

NLP research on artifacts has focused on (monolingual or mono-cultural) artifacts in texts, e.g., fairy tales, fiction, poetry and songs~\cite{yang-etal-2019-generating, haider-etal-2020-po, art/poetry/emnlp/Chakrabarty21, xu-etal-2022-fantastic, art/PAR3/emnlp/ThaiKKRIWI22, art/jiang-etal-2023-discourse, art/songs/ou-etal-2023-songs, art/songs/li-etal-2023-translate}, or in multimodal such as movies, humour and memes~\cite{sharma-etal-2020-semeval, liu-etal-2022-figmemes, hessel-etal-2023-androids, movie/tacl/hong-etal-2023-visual-writing}, to name a few.  While ``artifacts'' is an independent cultural element, usage in \textit{adaptation} typically involves tasks that align with one or more previously mentioned categories, making design challenging. For example, in ArtELingo (in \S\ref{sec:values}), the input data focuses on art, while cross-cultural measurement studies \textit{perceptions}, which reflect cultural values. Similarly, translations of literary novels need to account for \textit{concept} differences such as names~\cite{art/jiang-etal-2023-discourse} across cultures. Research on integrating cross-cultural differences into modelling and data acquisition with artifacts remains limited.

\subsection{Linguistic Elements}\label{sec:forms}

\subsubsection{Dialects} A dialect is a variant of a language~\cite{dialect/definition/haugen1966dialect} at the local regional level (e.g., Hessian German), national level (e.g., Tunisian Arabic) or by other factors (e.g., AAVE).

Many existing work focuses on dialect identification~\cite{dialect/salameh-etal-2018-fine,abdelali-etal-2021-qadi, dialect/hamalainen-etal-2021-finnish, dialect/id/yusuf-etal-2022-arabic}, but how to enable LLMs to serve dialectal communities remains an open question. 
Recently, multiple studies have identified disparities in NLP models \cite{dialect/value/acl/ziems-etal-2022-value, dialect/viet/emnlp/le-luu-2023-parallel, dialect/swiss/emnlp/paonessa-etal-2023-dialect, 
dialect/aave/emnlp/deas-etal-2023-evaluation} when evaluated across different language variations.

Current dialect datasets primarily consist of translations between dialects and standard languages or are created through dialect normalization, in text, audio, or both~\citep{dialect/swiss/acl/pluss-etal-2023-stt4sg, dialect/dialect2std/emnlp/kuparinen-etal-2023-dialect}. Few studies focus on traditional generation tasks like summarization or standard benchmark tasks (e.g., classifications or inferences \citealt{hate/xtremespeech/maronikolakis-etal-2022-listening, dialect/tada/acl/held-etal-2023-tada, dialect/dialectbench/faisal2024dialectbench}). Overall, research on German and English dialects is more advanced (marginally) than other dialect types.

\subsubsection{Styles, Registers and Genres}
Styles, registers (e.g., slang), and genres (e.g., news) depend on the context of language use \cite{lang_var/wardhaugh2021introduction}. 
Compared to other elements, recent developments in this area appear limited, with a handful of examples focusing on slang, formality or politeness \cite{style/emnlp/sun-xu-2022-tracing, formality/naacl/nadejde-etal-2022-cocoa, politness/emnlp/srinivasan-choi-2022-tydip, politness/emnlp/havaldar-etal-2023-comparing}.

\subsection{Social Elements}\label{sec:sociocult}

\subsubsection{Relationship} 
In many cultures, communication could differ depending on the relationship between the speakers. For example, Chinese has distinct terms for elder vs. younger siblings. Translations to (and from) a language without this property may result in a loss of nuances in meaning. In Korea and Japan, misused politeness level in conversation can violate cultural norms~\cite{matsumoto1988reexamination,ambady1996more}, especially in different social relationships. Additionally, certain relationships exist uniquely within specific cultures, such as ``Godmother''. Considering relationships is important for building resources and modelling culturally appropriate methods. \citet{nvm/renovi/naacl/abs-2402-11178} serve as a recent example with this consideration.

\subsubsection{Context} 
In NLP, linguistic context could be the surrounding text. Studies by \citet{hovy-etal-2020-sound, context/akinade-etal-2023-varepsilon, context/abs-2401-04972} show that machine translation systems can fail without appropriate consideration of linguistic context, revealing its importance in resource and model development. However, human communication is much richer, relying on the extra-linguistic context that situates language within broader frames of reference.

The extra-linguistic context can be situational (setting or location where communication occurs, e.g., at school, in a hospital), historical (past events, e.g., colonization, that change cultural values or language use, like in Hong Kong) or non-verbal (e.g., hand gesture, tone of voice). Each type shapes and reflects culture. These contexts significantly enhance conversational tasks, norm bank development, and visual-language applications~\cite{nvm/socialdial/sigir/zhan2023social, nvm/normbank/acl/ziems-etal-2023-normbank}, enabling NLP models to interpret nuanced language elements beyond words, thus improving response relevance and accuracy.

\subsubsection{Communicative Goals}
Different cultures can have distinctive communication styles depending on communicative goals. For example, people may use indirect language for refusal (versus direct refusal with a ``no'') to avoid confrontation~\cite{house2005politeness}. Cultures may also exhibit variations in responses to the same situation (e.g., how to make requests and when to apologize,~\citealt{blum1984requests}). Taking this type of variation into account is important for cross-cultural pragmatic-inspired tasks -- an area that remains understudied, with limited examples identified in Table \ref{tab:cult_elements}.

\subsubsection{Demographics} 
A household with a monthly income of less than 50 US dollars is likely to have different household items than that with 5000 US dollars~\cite{sociocultural/econ/dollarst/nips/RojasDKKRC22}. \citet{bias/frcrowspairs/acl/NeveolDBF22} also found that the original English CrowS-Pair dataset relied on names as proxies for a sociodemographic group (``Amy for women, Tyrone for African American men'', ~\citealt{bias/frcrowspairs/acl/NeveolDBF22}), whereas the French version features direct references to sociodemographic groups. These data differences may stem not only from cultural influences but also from the demographics of the data contributors. Where and from whom one collects data matters, as it can result in dramatic differences in data and modelling.

Demographic information is also important in annotation~\cite{demo/naacl/SapSVZCS22, demo/POPQUORN/pei-jurgens-2023-annotator, demo/acl/santy-etal-2023-nlpositionality}, where a piece of text can be humorous to some people but offensive to others~\cite{humour/meaney-2020-crossing}. In such cases, culture may exist in the labels rather than in the data. Recently, \citet{bias/crehate/abs-2308-16705} and \citet{hate/frenda-etal-2023-epic} show how to capture different cultural views of annotators using the same dataset. 

\subsection{Usage of the Taxonomy} 
Covering the key elements of culture, our taxonomy can act as a useful reference point for NLP system development, in addition to organizing existing literature. For example, the development of culturally aware debiasing should consider \emph{Ideational} elements such as \emph{Values}, \emph{Norms \& Morals} as well as \emph{Social} elements such as \emph{Demographics} to inform the focus of debiasing, along with \emph{Linguistic} elements such as dialects to inform the choice of data for the task. The applicable elements of culture will vary between tasks and contexts, with the taxonomy acting as a useful checklist.
See Appendix \ref{app:use} for additional examples.

\section{Culturally Aware Resource Acquisition}\label{sec:cult_data}

Resources discussed in \S\ref{sec:all_elements} are essential for culturally aware NLP. As additional resources are much needed, this section surveys methods for creating new resources (an overview in Appendix Figure~\ref{fig:overviews_data}).

Resources can be classified based on their acquisition methods—manual, automatic, or semi-automatic—and their source types: 1) newly created (\textbf{New}, from scratch), or 2) culturally adapted from existing resources (\textbf{CA}, e.g., through translation from the original data, followed by culturally appropriate changes). 
1) captures unique cultural phenomena but is often limited by funding or access to native speakers. 2) provides an alternative, though accurately reflecting cultural phenomena can be challenging.

\subsection{Manual: Incorporating Native Speakers, Communities, and Experts}\label{sec:manual_data}

A common strategy is to employ native speakers or experts (e.g., professional translators or students) for data acquisition. This can be done via crowd-sourcing platforms such as Amazon Mechanical Turk and Prolific~\cite{concept/marvl/emnlp/0001BPRCE21, concept/maps/naacl/liu2023} or in a community-driven manner, leveraging networks such as Masakhane~\footnote{\url{https://www.masakhane.io/}}, IndoNLP~\footnote{\url{https://indonlp.github.io/}}, university mailing lists, or Slack/Discord of organizations. Involving native speakers and communities to address cultural variations requires responsible design and thoughtful considerations. 

\textbf{New:} Most existing culture resources have been built by involving native speakers or communities for dataset acquisition \cite{concept/marvl/emnlp/0001BPRCE21, hate/xtremespeech/maronikolakis-etal-2022-listening, knowledge/indommlu/emnlp/KotoA0B23, concept/mabl/acl/KabraLKAWCAON23}. For non-language related communities, WinoQueer~\cite{bias/winoqueer/acl/FelknerCJM23} utilizes channels such as Slacks/Discord, and gay Twitter to reach the LGBTQ+ community and generates benchmarks based on community survey results.

\textbf{CA:} When starting from existing datasets, some works also involved communities (e.g., using surveys) in determining the needed modifications and supplements to datasets~\cite{bias/frcrowspairs/acl/NeveolDBF22, concept/multi3woz/tacl/Hu00609, concept/adapt/tacl/MajewskaRPVK23,  bias/kobbq/tacl/jin23}. These adaptations range from simple changes, such as updating names and locations to fit the target culture, to creating entirely new instances.

In general, native speakers are consulted throughout the life-cycle of new data acquisition (from annotations to quality checks). 
However, the entire community is rarely consulted during the initiation stage (i.e., designing tasks). Involving native speakers can be costly and difficult, but is a best practice that enhances quality and cultural authenticity.

\subsection{Automatic: Models and Pipelines}
Since manual adaptation is slow and hard to scale, the use of automation has gained popularity in resource acquisition.

\textbf{New: } For instance, CANDLE~\cite{kb/www/candle/NguyenRVW23} proposes a pipeline to extract cultural commonsense knowledge using various techniques like NER extraction, cultural facet classification, concepts extraction and ranking through algorithms or LMs. 
NormsSAGE~\cite{nvm/normsage/emnlp/fung-etal-2023-normsage} utilizes LLMs for norm discovery from conversation data, then performs model self-verification to validate and filter the data. 
CultureAtlas~\cite{kb/culturalatlas/fung2024massively} extracts cultural knowledge from Wikipedia and hyperlinked document pages using LLMs for filtering and adversarial knowledge generation. 

Recent works have also used sociodemographic prompting~\cite{demographic/icml/opinionsQA/SanturkarDLLLH23, demographic/deshpande-etal-2023-toxicity,demographic/hwang-etal-2023-aligning,demographic/eacl/BeckSLG24} --- extending input prompts with sociodemographic information --- to generate outputs tailored to specific groups. Further research could reduce data acquisition efforts, particularly for generating subcultural data variations within WEIRD people.
However, it has also been argued that 
LLMs do not accurately mimic individual or group behaviours~\cite{demographic/Argyle_Busby_Fulda_Gubler_Rytting_Wingate_2023, demographic/icml/AherAK23, demographic/eacl/BeckSLG24}. 

\textbf{CA: } 
\citet{knowledge/corr/abs-2402-17302} examine automatic adaptation (paraphrasing and concept replacement) of Commonsense QA in Indonesian and Sundanese. Current GPT models, however, reveal disparities in cultural adaptation across languages, highlighting the need for further research.

\subsection{Semi-Automatic: Structured Resources, Model-in-the-Loop}\label{sec:semi_data}

As demonstrated by \citet{knowledge/corr/abs-2402-17302}, LLMs struggle with fully automated cultural adaptations. Alternatively, semi-automatic approaches combine the quality of manual work with scalability.  

\textbf{New: } Methods have been developed to generate seed data for iterative human cleaning and labelling. NormBank~\cite{nvm/normbank/acl/ziems-etal-2023-normbank} uses LLMs to generate seed roles and behaviours as norm candidates in specific situations, which are then annotated by humans. Similarly, other studies \cite{nvm/SocNormNLI/emnlp/ch-wang-etal-2023-sociocultural, concept/maps/naacl/liu2023, bias/seegull/bhutani2024seegull} employ prompting techniques to generate seed data, followed by human annotation on tasks like cultural bias and social reasoning.

\textbf{CA:} \citet[Multi-Value]{dialect/multivalue/acl/ziems-etal-2023-multi} introduced a framework that leverages the Electronic World Atlas of Varieties of English~\cite[eWAVE]{ewave} to create and adapt datasets covering 50 English dialects. This framework enabled the adaptation of a standard corpus into dialectal forms~\citep{dialect/tada/acl/held-etal-2023-tada, adapt/transfer/emnlp/hyperlora}. However, similar structured resources may not exist or be suitable for adaptation of other cultural elements (e.g., for concepts, consistently replacing `bread' with `rice' would not be desirable). 

\section{Creating Culturally Adapted Models}\label{sec:cult_model}
Most culturally aware NLP research has focused on resource creation and evaluation, with culturally adapted model development still emerging. Here, we review current methods for adapting pre-trained (L)LMs, covering in-context and in-weight adaptations (an overview in Appendix Figure~\ref{fig:overviews_modles}). We found that current cultural adaptation methods in NLP prioritize technical advancements and isolated cultural elements, measuring effectiveness solely by standard task performance. 

\subsection{In-Context Adaptation}\label{sec:ki}
The success of LLMs allows for behaviour tuning by prompts or in-context examples.
A straightforward strategy is to provide the model with sociodemographic prompts or use ``role-playing'' \cite{roleplaying/conf/uist/ParkPCMLB22,demographic/Argyle_Busby_Fulda_Gubler_Rytting_Wingate_2023} of a culture, as seen in \citet{adapt/ki/codenames/shaikh-etal-2023-modeling} and \citet{demographic/hwang-etal-2023-aligning}. For knowledge-intensive tasks, cultural knowledge can be added directly to the prompt, and LLMs can leverage indirect descriptions from external sources or prior model outputs \cite{adapt/ki/translation/abs-2305-14328}. Lastly, high-level prompts (or ``constitutions'', \citealt{constituion_ai}) guiding LLM reasoning could improve cultural alignment alongside demographic-based prompts \cite{nvm/anthroprompt/acl/abs-2402-13231}.

Since different cultures reflect different values, there is a need to create models that embody pluralistic cultural values with flexible alignment capabilities \citep{pluralistic/icml/SorensenMFGMRYJ24}. \citet{adapt/modular/feng-etal-2024-modular} propose a framework to achieve this by enhancing pluralistic alignment in LLMs via collaboration between a high-level LLM and a group of specialized community LMs (i.e., an ensemble of LLMs). This framework enables general-purpose LLMs to flexibly incorporate diverse cultural and ideological perspectives, reflecting both individual preferences and broader cultural distributions.

A retrieval-augmented approach can further refine cultural alignment by adjusting responses dynamically. \citet{adapt/ki/values/ecai/FriedrichSSK23} propose such a method for moral reasoning, where culture-specific contexts are stored in a retrieval engine. When asked moral questions, relevant contexts are retrieved and added to the input, enabling the model to respond with cultural nuances. This method shows promise for adapting LLMs to evolving cultural information, an aspect often overlooked in current adaptation methods.

\subsection{In-Weight Adaptation}
\subsubsection{Data Augmentation}
Acquiring large corpora for supervised cultural adaptation is challenging. Data augmentation helps address this, enhancing model robustness.
\citet{adapt/dataaug/ccmm/li-zhang-2023-cultural} present a data augmentation method for multilingual multicultural VL reasoning tasks, generating code-mixed data by substituting English concepts with culturally mapped equivalents. The cultural concept sets (for mapping) are built by querying hyponyms, synonyms, and
hypernyms in the ConceptNet~\cite{conceptnet/aaai/SpeerCH17} and WordNet~\cite{wordnet/cacm/Miller95}. However, the optimal resource depends on the specific cultural element being adapted (\S\ref{sec:cult_elements}). For instance, a cultural knowledge base might be better for norms adaptations.

\subsubsection{Continual Pre-training, Auxiliary Losses}\label{sec:cpt}
Continual pre-training (CPT, including instruction tuning), intermediate task training, and multi-task training with auxiliary losses are methods for cultural adaptation. CPT fine-tunes a pre-trained LM with an unlabeled domain or language corpus before downstream task fine-tuning. It improves downstream task performance via full-parameter training~\cite{adapt/cpt/naacl/XuLSY19, adapt/cpt/emnlp/HanE19, adapt/cpt/acl/GururanganMSLBD20} or by training a few additional parameters while keeping the model frozen~\cite{adapt/cpt/acl/WangTDWHJCJZ21, adapt/cpt/emnlp/KeLS0SL22}.

Recently, \citet{adapt/cpt/tacl/hofmann22} show that when combined with a geo-location prediction loss, CPT can help to increase the awareness of dialectal variations of pre-trained LMs. \citet{cp/wang2024craft} show that instruction tuning with instructions containing cultural knowledge can improve models' ability in cultural knowledge reasoning. In VL, \citet{adapt/cpt/emnlp/bhatia-shwartz-2023-gd} use a cultural commonsense knowledge graph from \cite{kb/www/candle/NguyenRVW23} for CPT to develop a geo-diverse LM for commonsense reasoning tasks. This method category is effective for addressing diverse cultural elements, but adapting pre-trained LLMs can result in catastrophic forgetting (\citealt{MCCLOSKEY1989109}, or termed ``alignment tax'' due to RLHF tuning,~\citealt{alignmenttax/corr/abs-2112-00861, llm/nips/Ouyang0JAWMZASR22}) potentially worsening their performance on general tasks. This warrants further investigation.

\subsubsection{Other Forms of Information Integration}
% Missing cultural context, knowledge or demographic information have also been integrated into models in other ways:
% \sparagraph{As representations} 
\citet{adapt/ki/dialogue/eacl/CaoCH24} propose a method that integrates cultural dimension vectors (derived through a regression task based on Hofstede Culture Dimensions, \citealt{hofstede1984culture}) with a mT5 Transformer model \cite{mt5/naacl/XueCRKASBR21}. These cultural dimension vectors are added to the hidden states at each layer to enable culturally informed multi-turn dialogue classification and prediction.

\subsubsection{Parameter-Efficient Adaptations}\label{sec:peft}

As LMs grow larger, parameter-efficient fine-tuning methods (i.e., PEFT, by fine-tuning a small number of parameters, such as the bottle-neck adapters,~\citealt{DBLP:conf/icml/HoulsbyGJMLGAG19}; LoRA,~\citealt{DBLP:conf/iclr/HuSWALWWC22} etc.) become increasingly important for task adaptations. Given their success in cross-lingual transfer learning~\cite[among others]{DBLP:conf/emnlp/UstunBBN20, adapt/cpt/emnlp/pfeiffer-etal-2020-mad, DBLP:conf/emnlp/AnsellPPRGVK21, DBLP:conf/eacl/LiuPKVG23, fun}, PEFT can be a natural choice for cultural adaptation of e.g., dialects.  

Recently, HyperLoRA~\cite{adapt/transfer/emnlp/hyperlora} uses the Hypernetwork~\cite[a neural network for generating parameters]{DBLP:conf/iclr/HaDL17} to generate LoRA adapters based on dialectal features. DADA~\cite{adapt/transfer/emnlp/liu-etal-2023-dada} proposes to train a pool of dialectal linguistic feature adapters and dynamically compose the adapters for dialectal tasks. 
Being task agnostic, PEFT methods could prove important for cultural adaptations beyond dialects.

\subsubsection{Outlook: Feedback Learning}\label{sec:dpo}

The success of LLMs has popularized Reinforcement Learning from
Human Feedback~\cite[RLHF]{rlhf/nips/ChristianoLBMLA17,rlhf/corr/abs-2204-05862, llm/nips/Ouyang0JAWMZASR22, feedbacklearning/nips/IvisonW0WP0S0H24} and Direct Preference Optimization~\cite[DPO]{dpo/nips/RafailovSMMEF23,dpo/corr/abs-2311-10702} methods. RLHF fine-tunes LMs with feedback by fitting a reward model with human preferences, and then training a reinforcement learning-based policy to maximize the learned reward. DPO avoids RL training by using a simpler supervised learning objective for an implicit reward model.

Recent work shows that RLHF can enhance the performance of multilingual instruction tuning for LLMs~\cite{rlhf/emnlp/LaiNNNDRN23}, while DPO can improve the multilingual reasoning abilities~\cite{dpo/mapo/abs-2401-06838} and multilingual safety \cite{adapt/rlhf/arash-etal-2024-multilingual} of LLMs. The use of RLHF or DPO for multilingual multicultural adaptation is still limited, but these examples suggest that the direction could be promising.

\section{Further Discussions and Recommendations}\label{sec:discussion}
As we have seen, significant work remains to be done on both resources and methods for various elements of culture.

An area that requires attention is the overall process of researching culturally aware NLP. As mentioned previously, a key practice is community involvement (\S\ref{sec:cult_data}) to get the process right~\cite{indigenous/bird-2020-decolonising, indigenous/liu-etal-2022-always, indigenous/mager-etal-2023-ethical}. It is crucial to assess how target communities can benefit most from technologies. For instance, many dialects are primarily oral, and speech-to-speech or speech-to-text translations could be preferable over text-based applications~\cite{dialect/survey/blaschke2024dialect}. Furthermore, ethical data collection practices are also critical and technology ownership must be considered, especially when indigenous and marginalized communities are involved.  For best practices, we refer the readers to work such as~\citet{indigenous/bird-2020-decolonising}, \citet{smith2021decolonizing} or \citet{indigenous/cooper-etal-2024-things} for further details. 

Another key consideration is integrating insights from fields beyond NLP. 
Cultural adaptation has long been practiced in areas like video games~\cite{o2013game}, movies~\cite{pettit2009connecting}, online learning~\cite{blanchard2005cross}, and clinical psychology~\cite{evm/bernal1995ecological, cp/barrera2006heuristic}. These existing practices can serve as a foundation for adapting NLP applications to meet the needs of diverse cultural contexts. 

Here, we summarize and recommend best practices based on our prior discussions and the publications surveyed in the sections referenced below:

\noindent{\textbf{Resource Acquisition.}} 
\begin{itemize}[noitemsep, topsep=0pt]
    \item \S\ref{sec:manual_data}, \S\ref{sec:discussion}: Consult with target cultural groups throughout design and implementation, wherever possible. 
    \item \S\ref{sec:semi_data}: Use iterative feedback from culture experts to refine data quality.  Automatically acquired resources should also undergo expert quality checks.
    \item \S\ref{sec:discussion}: Ensure an ethical approach to data acquisition and discuss data ownership early to prevent misuse. This is always important but particularly critical with indigenous and marginalized communities.
\end{itemize}

\noindent{\textbf{Model Adaptation.}} 
\begin{itemize}[noitemsep, topsep=0pt]
    \item \S\ref{sec:cult_model}: Incorporate new metrics that assess cultural awareness alongside task performance.
    \item \S\ref{sec:cult_model}: Consider cultural adaptation as an ongoing, systematic process rather than a one-time task focused on a single element.
    \item \S\ref{sec:ki}: Monitor adaptation performance over time, especially for the evolving cultural elements, to maintain model relevance.
    \item \S\ref{sec:discussion}: Build on existing knowledge outside of NLP when applicable. 
\end{itemize}

\section{Summary and Future Research Directions}\label{sec:gaps}

Culturally aware and adapted NLP has recently emerged as an important and active research area. Significant progress has been made in the development of resources for capturing various elements of culture, but the development of NLP methods is still in its infancy. 
We will now summarize the main research gaps identified in this survey with respect to the categories of our new taxonomy (\S\ref{sec:framwork}):

\sparagraph{Resources} Currently, resources exist for all elements of culture, with considerable progress made on \textit{values} (\S\ref{sec:values}, particularly in biases) and \textit{knowledge} (\S\ref{sec:knoweldge}, particularly for MMLU-style cultural knowledge benchmarks). However, research is lacking in the following areas:

\itparagraph{Gaps in Elements Coverage} While many resources already exist within \emph{concepts} (\S\ref{sec:concept}), multilingual data resources covering a diverse set of concepts (e.g., aesthetics, spatial relation) in both unimodal and multimodal (\S\ref{sec:concept}) for generation tasks is lacking. Moreover, most recent developments in \textit{norms \& morals} are predominantly in English, reflecting a monocultural perspective. This highlights the need for more multilingual and multicultural resources. Additionally, there is a significant gap in datasets that focus on different types of value perceptions (such as emotion and irony, \S\ref{sec:values}), stylistic variations (\S\ref{sec:forms}), and artifacts (\S\ref{sec:artifacts}) across various cultural groups, both in different languages and within languages. 

Resources considering social elements of culture (\S\ref{sec:sociocult}) also remain limited. For example, collecting speaker relationships in dialogue datasets or distinguishing age groups in social norms datasets. These are needed to address the intricate relationship between culture and people in NLP. 

\itparagraph{Training Data and ``CultureGLUE''} 
Most existing resources focus on evaluation, providing benchmarks and test sets that enable researchers to assess the performance of models. While these evaluation resources are crucial for cultural adaptation, there is a pressing need for training data. Further, current evaluation resources often focus on individual elements of culture. A unified, cultural benchmark like GLUE \cite{wang-etal-2018-glue} does not yet exist for all cultural elements across diverse groups. Developing a multicultural ``CultureGLUE'' may be challenging at the moment, but a reasonable first step is to focus on individual cultures, ensuring a diverse range of tasks and comprehensive element coverage.

\sparagraph{Modelling} While modelling methods for culture are generally under-explored, continual pretraining (\S\ref{sec:cpt}) and prompting (\S\ref{sec:ki}) have received marginally more attention than other approaches. 
Research areas needing further exploration include:

\itparagraph{PEFT-based Transfer Learning} Exploration of PEFT-based transfer learning techniques beyond dialects is limited (\S\ref{sec:peft}). Given their success in other NLP areas, these techniques warrant further investigation into other elements of culture, such as for \textit{values} or \textit{norms \& morals}. A potential approach involves using WVS survey data, similar to \citet{cultureLLM/corr/abs-2402-10946}, to train PEFT-based modules focused on values. However, it is crucial to investigate whether survey data alone is sufficient for effective training.

\itparagraph{Feedback Learning and Other LLM Specialties} 
Leveraging the success of LLMs and feedback learning presents promising new avenues for cultural adaptation (\S\ref{sec:dpo}). A potential bottleneck is acquiring large, culturally diverse preference datasets for model adaptation and training culturally aligned reward models. This could be addressed by large-scale data collection efforts, as demonstrated by \citet{nvm/kirk2024prism}, or through the generation of synthetic data \cite{adapt/rlhf/arash-etal-2024-multilingual}. For synthetic data, using techniques such as role-playing and the creation of repositories of cultural personas could facilitate culturally sensitive model training.

\itparagraph{Evolving Culture} Culture evolves gradually~\cite{boyd1988culture, whiten2011culture}, yet there have been few discussions on how to model and adapt to evolving culture. Future research should focus on methods that address the dynamic nature of culture.
One potential approach is the use of retrieval-augmented systems to integrate evolving information (\S\ref{sec:ki}), which ensures models' relevance to cultural shifts over time. 

\sparagraph{Overall} Below, we discuss two overall research gaps.

\itparagraph{Adaptation in the Social Context} As a key motivation of this paper, culture emerges from and is shaped by social interactions among humans within a society (\S\ref{sec:intro}). However, an important question remains unanswered in the existing literature: \emph{Should the cultural adaptation of models occur within a situated social context and structure?} Exploring this could present new avenues for interdisciplinary research (e.g., with human-machine collaboration, social psychology, or anthropology etc.).

\itparagraph{``Surface'' versus ``Deep'' Adaptation in NLP}
% \sparagraph{Overall - ``Surface'' versus ``Deep'' Adaptation in NLP}
\citet{cp/csm/42269711-6e5b-382a-b370-098ad7128da0} devise cultural adaptations for public health research into surface and deep adaptations, where the former considers familiar languages and concepts to the target groups, and the latter considers social and historical factors that influence the behaviours of the target groups. 

In NLP, surface adaptations might include using the same language as a culture and recognizing explicit cultural differences (e.g., asking LLMs ``what is the meaning of ...''). In contrast, deep adaptations might enable a model to ``behave'' (e.g., make decisions, pragmatically comply etc.) like a member of a culture without explicit inquisition (see Figure~\ref{fig:surface_vs_deep} in the Appendix for an illustration).

As we have seen from prior work described in \S\ref{sec:all_elements} or \S\ref{sec:cult_model}, only a few current works focus on adapting the behavioural aspect of models (which is becoming increasingly important with LLMs), and there has been no work to date on measuring the depth and progress of cultural adaptations or when a model is \textit{fully culturally aware and culturally competent}. Further research could explore these areas.

\section{Conclusions}

This work proposes a new extensive taxonomy of culture that expands on earlier works in NLP and is grounded in well-established anthropology and social sciences literature. The taxonomy provides a systematic framework for understanding and tracking progress in the emerging area of culturally aware and adapted NLP. However, our taxonomy is not without its limitations. Future research could refine the taxonomy in areas like \textit{values} or \textit{communicative goals} by adding further subcategories and providing a better understanding of interactions between the elements of culture (e.g., shifting values' impact on social norms over time).

We survey existing resources and methods in this area according to the taxonomy classes, identifying areas of strength as well as areas where research remains to be done. Our paper summarizes the state of the art and provides ideas for future research in this exciting and important area. 

\section*{Acknowledgements}
This work was funded by the German Federal Ministry of Education and Research (BMBF) under the promotional reference 13N15897 (MISRIK). It was also funded by EPSRC grant EP/Y031350/1 under the UK government’s funding guarantee for ERC Advanced Grants. Chen Cecilia Liu is supported by the Konrad Zuse School of Excellence in Learning and Intelligent Systems (ELIZA) through the DAAD program Konrad Zuse Schools of Excellence in Artificial Intelligence, sponsored by the Federal Ministry of Education and Research.

The authors would like to thank Anjali Kantharuban, Shun Shao, and Anne Lauscher for an early discussion of different aspects of this work. The authors would also like to thank Ji-Ung Lee, anonymous reviewers and action editors of TACL for feedback on a draft of this paper.

\clearpage
\newpage
\bibliography{culture_revised}
\bibliographystyle{acl_natbib}

\clearpage
\appendix

\section{Method}\label{app:method}
We examine the main and findings papers from the leading *CL venues, including: ACL, EMNLP, AACL, EACL, NAACL and TACL published since 2020 (5-year span). The initial set of papers was identified using the following search terms: ``culture'', ``cultural'', ``geo-diverse'', ``socio'', ``social'', ``moral'', ``norms'' in the title and abstract. Initially, we collected 336 papers, using human verification to exclude papers that did not consider cultural variations, as well as papers that solely focus on analysis and probing (as they are beyond the scope of our survey). The final paper count is 127. For more on probing and analysis, please refer to the recent surveys like \citet{DBLP:journals/corr/abs-2403-15412}. We further acknowledge the limitation of missing relevant papers from other sources and papers without explicitly mentioning any of the search keywords. However, our goal is not to conduct a systematic review, but to propose a taxonomy and understand the progress in NLP for this research area and identify research gaps. We believe that focusing on *CL venues is an appropriate choice for this purpose.

\section{Additional Examples of Use Cases for the Taxonomy.}\label{app:use}

Another example of applying this taxonomy is the development of culturally aware conversational AI for educational purposes. Such development should be informed, at a minimum, by relevant \emph{Knowledge} (e.g., Facts), appropriate \emph{Style}, understanding of the \emph{Communicative Goals} (e.g.,  that of teaching) and consideration of \emph{Relationships} (e.g., that of a teacher and student). These are merely example elements and applications to consider.

\begin{figure}[h!]
\vspace{10pt}
     \centering
     \includegraphics[width=\linewidth]{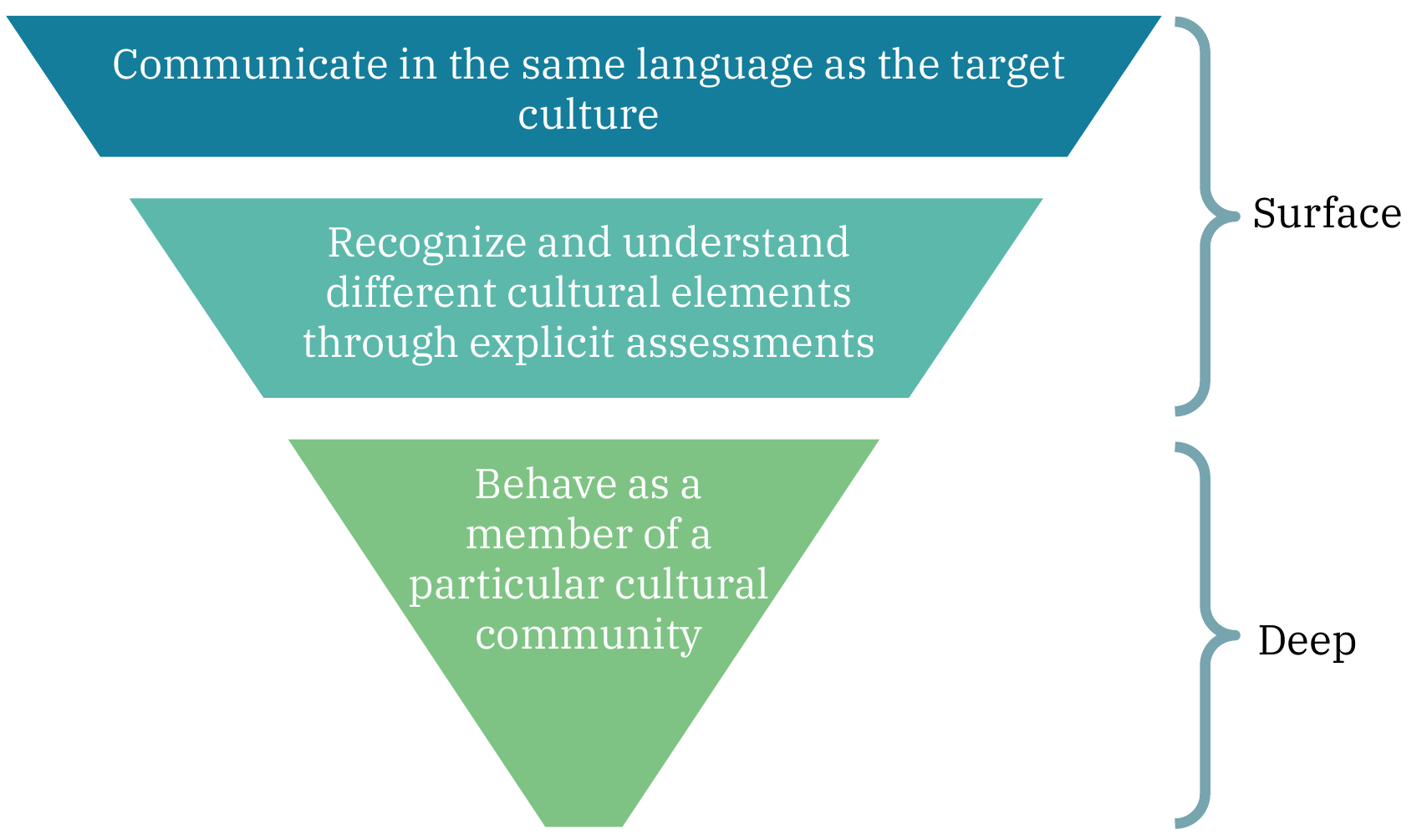}
     \caption{An illustration of surface versus deep culturally adapted NLP model.}
     \label{fig:surface_vs_deep}
 \end{figure}

\tikzset{%
    parent/.style =          {align=center,text width=2cm,rounded corners=3pt, line width=0.3mm, fill=gray!10,draw=gray!80},
    child/.style =           {align=center,text width=2.3cm,rounded corners=3pt, fill=blue!10,draw=blue!80,line width=0.3mm},
    grandchild/.style =      {align=center,text width=2cm,rounded corners=3pt},
    greatgrandchild/.style = {align=center,text width=1.5cm,rounded corners=3pt},
    greatgrandchild2/.style = {align=center,text width=1.5cm,rounded corners=3pt},    
    referenceblock/.style =  {align=center,text width=1.5cm,rounded corners=2pt},
    data_pt/.style = {align=center,text width=2cm,rounded corners=3pt, fill=paired-light-green!50,draw=paired-dark-green!65,line width=0.3mm},   
    data/.style =           {align=center,text width=2.5cm,rounded corners=3pt, fill=paired-light-green!50,draw=paired-dark-green!65,line width=0.3mm},   
    data_sub/.style =           {align=center,text width=1.5cm,rounded corners=3pt, fill=paired-light-green!50,draw=paired-dark-green!65,line width=0.3mm},   
    data_work/.style =      {align=left, text width=3.2cm,rounded corners=3pt, fill=paired-light-green!50,draw=paired-dark-green!80,line width=0.3mm} 
}

\begin{figure*}[]
    \scriptsize
    \centering
    \resizebox{0.9\textwidth}{!}{
    \begin{forest}
        for tree={
            forked edges,
            grow'=0,
            draw,
            rounded corners,
            node options={align=center,},
            text width=2cm,
            s sep=4pt,
            calign=child edge, 
            calign child=(n_children()+1)/2,
            l sep=7.5pt,
        },
        [, phantom
            [Data \S\ref{sec:cult_data}, data_pt
                [Manual, 
                data[New, data_sub[
                                \citet{concept/marvl/emnlp/0001BPRCE21}\\
                                \citet{concept/xm3600/emnlp/ThapliyalPCS22}\\
                                \citet{bias/winoqueer/acl/FelknerCJM23}\\
                                \citet{knowledge/indocult/koto2024indoculture}\\
                                \citet{bias/crehate/abs-2308-16705}, data_work]]
                data[Adapt, data_sub[
                \citet{concept/adapt/tacl/MajewskaRPVK23}\\
                 \citet{knowledge/indommlu/emnlp/KotoA0B23}\\
                \citet{concept/multi3woz/tacl/Hu00609}, data_work] ]
                ]
                [Automatic: Models and Pipelines,  
                data[New, data_sub[\citet{kb/www/candle/NguyenRVW23}\\
                            \citet{nvm/normsage/emnlp/fung-etal-2023-normsage}\\
                            \citet{kb/culturalatlas/fung2024massively}, data_work]]
                data[Adapt, data_sub[\citet{knowledge/corr/abs-2402-17302}, data_work]]
                ]
                [Semi-Automatic: Model-in-the-loop; Structured Resources,
                data[New, data_sub[\citet{nvm/normbank/acl/ziems-etal-2023-normbank}\\
                \citet{nvm/SocNormNLI/emnlp/ch-wang-etal-2023-sociocultural}\\
                \citet{concept/maps/naacl/liu2023}\\
                \citet{bias/seegull/bhutani2024seegull}
                , data_work]]
                data[Adapt, data_sub[
                \citet{dialect/tada/acl/held-etal-2023-tada}\\
                \citet{adapt/transfer/emnlp/hyperlora}
                , data_work]]
                ]
            ]
        ]
    \end{forest}
    }
    \caption{Categorization of the methods for resource acquisitions with representative examples.}
    \label{fig:overviews_data}
\end{figure*}

\tikzset{%
    parent/.style =          {align=center,text width=2.3cm,rounded corners=3pt, line width=0.3mm, fill=gray!10,draw=gray!80},
    child/.style =           {align=center,text width=2.3cm,rounded corners=3pt, fill=blue!10,draw=blue!80,line width=0.3mm},
    grandchild/.style =      {align=center,text width=2cm,rounded corners=3pt},
    greatgrandchild/.style = {align=center,text width=1.5cm,rounded corners=3pt},
    greatgrandchild2/.style = {align=center,text width=1.5cm,rounded corners=3pt},    
    referenceblock/.style =  {align=center,text width=1.5cm,rounded corners=2pt},
    finetuning_pt/.style =           {align=center,text width=2cm,rounded corners=3pt, fill= paired-light-red!50,draw=paired-dark-red!70,line width=0.3mm}, 
    finetuning/.style =           {align=center,text width=3cm,rounded corners=3pt, fill= paired-light-red!50,draw=paired-dark-red!70,line width=0.3mm}, 
    finetuning_work/.style =      {align=left,text width=3.2cm,rounded corners=3pt, fill= paired-light-red!50,draw= paired-dark-red!90,line width=0.3mm}, 
}

% !htb
\begin{figure*}[]
    \scriptsize
    \centering
    \resizebox{0.9\textwidth}{!}{%
    % \begin{minipage}[b]{0.85\linewidth}
    \begin{forest}
        for tree={
            forked edges,
            grow'=0,
            draw,
            rounded corners,
            node options={align=center,},
            text width=2cm,
            s sep=4pt,
            calign=edge midpoint,
            l sep=7.5pt,
        },
        [, phantom
            [Modelling \S\ref{sec:cult_model}, finetuning_pt
            [In-Context, finetuning_pt
            [Sociodemographics Prompt; Cultural Role-playing, finetuning[
            \citet{adapt/ki/codenames/shaikh-etal-2023-modeling}\\
            \citet{demographic/hwang-etal-2023-aligning}\\
            \citet{nvm/anthroprompt/acl/abs-2402-13231}
            , finetuning_work]
            ]
            [Cultural Knowledge, finetuning[
            \citet{adapt/ki/translation/abs-2305-14328}\\
            \citet{knowledge/FMLAMA/zhou2024does}\\
            , finetuning_work]
            ]
            [Reasoning Prompt; \\Constitution, finetuning[
            \citet{adapt/prompting/emnlp/hayati2024}\\
            \citet{nvm/anthroprompt/acl/abs-2402-13231}\\
            \citet{adapt/ki/vijjini-etal-2024-socialgaze}\\
            , finetuning_work]
            ]
            [Ensemble of LLMs, finetuning[
            \citet{adapt/modular/feng-etal-2024-modular}\\
            , finetuning_work]
            ]
            [Retrieval-Augmented, finetuning[
            \citet{adapt/ki/values/ecai/FriedrichSSK23}\\
            \citet{adapt/ki/emnlp/conia-etal-2024-towards}\\
            \citet{adapt/retrieval/emnlp/hu-etal-2024-bridging}
            , finetuning_work]
            ]
            ]
            [In-Weight, finetuning_pt
                [Data Augmentation, finetuning
                [\citet{adapt/dataaug/ccmm/li-zhang-2023-cultural}, finetuning_work]
                ]
                [Continual Pre-training; Auxilary Losses, finetuning [
                \citet{adapt/cpt/emnlp/bhatia-shwartz-2023-gd}\\
                \citet{hate/cultcompass/emnlp/ZhouKCH23}\\
                \citet{adapt/cpt/tacl/hofmann22}\\
                \citet{adapt/mtl/dialect/spliethover-etal-2024-disentangling}\\
                \citet{cp/wang2024craft}\\
                \citet{adapt/cp/acl/cahyawijaya-etal-2024-cendol}\\
                \citet{nvm/acl/kim-etal-2024-moralemotion}\\
                \citet{cda/data/cidar/alyafeai-etal-2024-cidar}\\
                , finetuning_work]
                ]
                [Information Integration (with vectors), finetuning
                    [
                    % \citet{adapt/ki/values/ecai/FriedrichSSK23}\\
                    \citet{adapt/ki/dialogue/eacl/CaoCH24}\\
                    \citet{adapt/ki/codenames/white-etal-2024-communicate}
                    , finetuning_work]
                ]
                [Parameter-Efficient Adaptations, finetuning
                    [\citet{adapt/transfer/emnlp/hyperlora}\\
                    \citet{adapt/transfer/emnlp/liu-etal-2023-dada}, finetuning_work]
                ]
                [Feedback Learning, finetuning
                    [\citet{adapt/rlhf/arash-etal-2024-multilingual}
                    , finetuning_work]
                ]
            ]
            ]
        ]
    \end{forest}
    }
    \caption{Categorization of the adaptation modelling methods and examples in each category.}
    \label{fig:overviews_modles}
\end{figure*}

\end{document}